# A roadmap for generative mapping: unlocking the power of generative AI for map-making

Sidi Wu*
sidiwu@ethz.ch
Institute of Cartography and Geoinformation, ETH Zurich
Switzerland

Katharina Henggeler*
khenggel@ethz.ch
Institute of Cartography and Geoinformation, ETH Zurich
Switzerland

Yizi Chen*
yizi.chen@ethz.ch
Institute of Cartography and Geoinformation, ETH Zurich
Switzerland

Lorenz Hurni
lhurni@ethz.ch
Institute of Cartography and Geoinformation, ETH Zurich
Switzerland

## ABSTRACT
Maps are broadly relevant across various fields, serving as valuable tools for presenting spatial phenomena and communicating spatial knowledge. However, map-making is still largely confined to those with expertise in GIS and cartography due to the specialized software and complex workflow involved, from data processing to visualization. While generative AI has recently demonstrated its remarkable capability in creating various types of content and its wide accessibility to the general public, its potential in generating maps is yet to be fully realized. This paper highlights the key applications of generative AI in map-making, summarizes recent advancements in generative AI, identifies the specific technologies required and the challenges of using current methods, and provides a roadmap for developing a generative mapping system (GMS) to make map-making more accessible.

## 1 INTRODUCTION

Maps, as symbolic representations of space and spatial knowledge, serve as useful tools for visualizing, analyzing, and communicating spatial information. Map-making is relevant not just for cartographers but for various professionals and even laymen. For example, spatial planners make maps to communicate their designs and record decisions as well as use map-making as a research method during participatory approaches. Journalists and publishers use maps to present spatial topics and results, complementing the textual materials. Teachers use maps to illustrate and explain various regions' geography, history, and culture. The tourism industry employs maps to guide travelers and showcase attractions.

Thanks to the advent of GIS software and platforms (e.g., ArcGIS, QGIS), map-making has become significantly more accessible. Despite these advances, map-making still requires expertise in cartography and GIS. Also, the whole workflow can be time-consuming, particularly for well-designed maps that show complex topics. As a result, people pressed for time or without cartographic and GIS knowledge, as well as the public in general—while heavily using maps (e.g., for navigation or as a source of knowledge)—do not have the means to create maps themselves. For that to change, map-making would need to be as easy as taking a picture: with a command as simple as "create a black-and-white world map showing only continents and no names".

This is exactly the opportunity that generative AI suggests. Unlike discriminative models that classify or predict outcomes based on existing data, generative models learn the data distribution to create new data. This is essential for capturing the "design space" of maps and enabling users to give instructions (prompts) to customize maps according to specific preferences and purposes. While generative AI models have been used for tasks such as accurate mapping from satellite images [31], the potential for map customization that meets diverse user needs using multi-modal prompts has not yet been fully explored.

This paper provides a comprehensive overview of key aspects and techniques required for generative AI in map generation and provides a roadmap for developing **generative mapping systems (GMSs)** considering various use cases, distinguishing itself from prior research that focused on the ethical drawbacks [17] and user perspectives in AI-driven cartography [24].

## 2 APPLICATIONS OF GENERATIVE AI FOR MAP-MAKING

We summarize the key applications of generative AI in map-making as follows:

**Automatic, rapid, and accessible map generation.** Given the user's specified inputs (e.g. texts, images, sketches, or geodata), the generative AI models can automatically generate maps rapidly without requiring much GIS or cartographic knowledge. For example, a teacher can easily generate "a black-and-white world map showing only continents without place names", or a disaster management office can instantly generate maps updated from satellite images.

**Customized maps.** Anyone can use generative AI models to customize maps by the specified requirements like map types (e.g., topographic, thematic, pictorial), styles, scales, and editing suggestions (e.g., incorporating, highlighting, or deleting certain objects).

---
*Equal contribution.

**Spatial simulation.** Maps can be generated from descriptions or sketches to simulate scenarios under various hypotheses, aiding decision-making. Possible applications lie in research, policy-making, and planning. For example, an urban planner may ask, "How would this region look like if no railways had been built?"

**AI-assisted designs.** The generative AI models can assist the map-designing process by providing various potential designing schemes to encourage more creative designs.

**Data synthesis and augmentation.** The generative AI models allow the synthesis and augmentation of map data to support machine learning models. For example, modern vector data can be synthesized into historical map styles to bootstrap the information extraction from historical maps.

## 3 ADVANCES AND CHALLENGES OF GENERATIVE AI IN MAPS

### 3.1 Advances in Generative AI

*Generative AI* is a class of artificial intelligence aiming at generating new and meaningful content that resembles the patterns and distributions of the original training data [7]. Each data sample in the distributions learned from the generative model represents a unique and distinct representation, resembling the original dataset used to train the generative models. Common generative models are Generative Adversarial Networks (GANs) [10, 33], variational autoencoders (VAEs) [19, 20], autoregressive models [29], and diffusion models [5]. The generative models can generate various forms of data in different applications, including creating text content, generating realistic images and videos [13, 14], and even synthesizing human-like speech [15, 16]. Generating maps can be comparable to generating images as both maps and images are graphical representations, which makes generation models potentially transferable. Recent works have shown great advances in generating photo-realistic images [27, 28] and further control the quality of generated images by incorporating textual guidance [2, 3, 8, 18] and image guidance such as segmentation [1, 9, 25], sketches [30] or scribbles [32].

### 3.2 Challenges of using generative AI in maps

Despite the similarity between image and map generation, they also have distinct differences. Images capture visual information of reality. Thus, image generation aims to generate visually realistic images with high diversity. On the other hand, maps are symbolized and abstracted representations of a (geographical) reality. They are used to illustrate spatial relationships [26, 21]. Thus, they aim for topological accuracy and preserving geometrical properties such as shape, area, or direction. Also, maps can have different scales with different levels of detail besides different resolutions. Therefore, a map generation model should suitably abstract, generalize, and represent the data at different scales with maximum topological accuracy and plausible geometry. Current image generation models are insufficient in fulfilling these goals.

Moreover, image generation models usually produce only small-sized results (e.g. 512*512 pixels). These are insufficient for large-scale maps with fine-grained details. Distribution-based metrics (like FID, KID) and pixel-based metrics (like RMSE) are used to evaluate the quality of generated images, whereas metrics to assess the spatial relationships, topology, and geometry remain untapped. Besides raster maps, it's also important to generate vector maps containing more complex geometry (points, lines, and polygons) of various sizes compared with the simple vector graphics of SVG icons researched in [4].

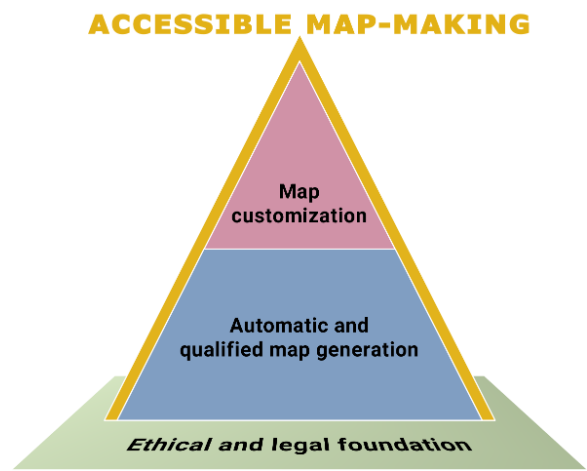

**Figure 1: Roadmap towards accessible map-making with generative AI.**

While current generative AI models exhibit promising capabilities in interpreting text prompts, texts alone can be imprecise and insufficient in describing map elements (e.g., spatial relationships, distances of objects, and spatial layout) and ambiguous depending on the user's expertise level and cultural background. This may require additional inputs (e.g., sketches, images, and other geodata sources) or smarter interactions with the generation process. Also, various geodata inputs from different sources present a large diversity in structure, dimension, resolution, granularity, and accuracy. Nevertheless, existing generative AI models focus more on limited modalities without considering the gap or misalignment between different sources.

## 4 OPEN QUESTIONS AND RESEARCH OPPORTUNITIES

Figure 1 illustrates a roadmap for developing GMSs. To achieve the ultimate goal of making map-making accessible to everyone, the system should generate qualified maps automatically and rapidly as well as allow customization independent of the user's cartographic and GIS knowledge. The system should be contextualized by an ethical and legal foundation that recognizes national and cultural differences. In this section, we present the roadmap in detail by highlighting open questions and research opportunities.

### 4.1 Automatic and qualified map generation

The base of a GMS is the automatic and qualified generation of maps. This requires a model capable of making suitable map design choices. More precisely, it must:

**Initialize and generate qualified maps.** The basic functionality of GMS is to generate maps with plausible geometry and topology as well as correct spatial relationships. It should initialize such maps in the first place to avoid rounds of follow-up prompts for improving the map quality, which is inefficient and a sign of poor usability. This means that explicit topology and geometry control should be emphasized to guide the training process. Instead of diverse yet incorrect results, the model should narrow the possibilities down to the most common and suitable representations (including map types, symbols, and scales) following cartographic rules. For example, "a map of Switzerland showing population densities" will lead to the model to automatically generating a choropleth map showing the statistics at cantonal level by default.

**Process and combine (mismatched) multi-modal data.** Not only should the model be able to process multi-modal data like statistical files, satellite images, texts, street-view images, and vector data [22]; the model should also be able to link and combine the data on a semantic level. Besides, different data sources can have mismatches due to different capturing methods, resolutions, and acquisition times. For example, modern vector data has more dense objects and finer granularity of object categories than historical sources of vector data. The mismatch between data sources should be explicitly considered in the map generation process.

**Avoid fake and misleading toponyms.** Maps typically contain toponyms or place names. Existing generation models tend to hallucinate unmeaningful toponyms to mimic the data distribution [31], which reduces the usability of generated maps. Therefore, it is crucial to detect and filter out these fake and misleading toponyms. The approach in [31] identified toponyms as mismatches between satellite images and topographic maps and prevented the GAN model from including such information in the generated maps. It serves as an effective basis to eliminate toponyms in the generation, which can be further included in other generation models (like diffusion). Accurate toponyms can subsequently be superimposed onto the generated, clean maps.

**Abstract information and generalize.** Maps of different scales should contain different levels of detail, requiring the model to establish a hierarchy that represents various levels of abstraction, either based on content (e.g. categories, attributes) or geometry (e.g. adjacency, proximity, alignment, intersection). When the scale decreases, the model should reduce details, aggregate information, and eliminate less important information. Although deep learning models have been developed for automating topographic map generalization [12], there is still limited research on knowledge abstraction for thematic maps. This might be enhanced by integrating large language models (LLMs) to summarize the information and build the knowledge graphs [23].

**Generate scalable and coherent outputs at different locations.** Image generation models usually only support small-sized outputs. Therefore, tiling techniques are needed to generate and stitch small map-tiles (images) into the composite large-scale map. This process should be aware of the context to ensure coherence in the abstraction and generalization levels, map styles, and geometry across different regions. This may require an iterative generation treating all the tiles as a sequence of results based on the previous result [6].

**Output the map in an appropriate format.** Graphical formats such as JPEG, PNG, and SVG are acceptable in many use cases (e.g., for authors, journalists, and school teachers). However, the model should also be able to output geospatial formats like GeoTiff or Shapefile with any coordinate system of choice to support further analysis or editing. Besides these static 2D output formats, more complex forms like dynamic, interactive, and 3D formats should be further considered.

## 4.2 Map customization

Built on top of 4.1, customisation functionality should be provided in order to adjust the maps according to the user's design choices. To this end, the system should:

**Understand and link map content and instructions.** The model needs to understand the map contents it has created and will create, and correlate the user's instructions (e.g., text prompts) to the content, in order to correctly react to the customization requests. For example, "shift the river 500m north" would first require the model to understand the semantics from the base map and link the text prompt to the referred object in the map. A map foundation model should be developed that bridges the gaps between vision (maps) and language (prompts).

**Offer different interaction modalities.** Different interaction modalities besides the standard written text prompt can be imagined for GMSs. Users could draw on a generated map to indicate desired adjustments. Similarly, one could allow simple ways of modifying map objects (e.g., adding, removing, moving, and scaling). Generally, the more complex the offered interaction modalities, the more cartographic knowledge is needed to use them and the closer the model will come to GIS software. Thus, interaction modalities should be kept simple in order to maximise accessibility.

**Handle non-expert and ambiguous prompts.** The model should handle ambiguous and non-expert prompts by linking them to existing GIS and cartographic concepts or by requesting clarification, enabling accessibility for users across the full knowledge spectrum of GIS and cartography. For example, "maps with buildings and rivers" likely refer to topographic maps; "can you make the map coarser" would refer to map generalization and the system can ask for the desired level of coarseness (the generalization level). This may require fine-tuning LLMs to bridge the gap between non-expert language and special GIS and cartographic concepts.

## 4.3 Ethical and legal foundation

As mentioned in 4.1, the GMS should be able to generate qualified maps. While there can be objective metrics measuring the quality of geometry and spatial relationships of generated maps, the criteria for a qualified map can vary by, for example, cultural, social and legal contexts, making it challenging to establish a one-size-fits-all standard for map quality. Therefore, the GMS must be adaptable and sensitive to these differences to generate maps that are considered qualified and acceptable in diverse contexts, which the model can allow or request the user to specify. Maps shape our perceptions and decisions but are also inherently biased [11]. Training an unbiased model based on existing maps is therefore nearly impossible. This can potentially be mitigated by identifying biases, especially for highly disputable issues, and treating them as contextual prompts to control the model outputs. These contextual prompts can be further extended to cultural and legal conventions of map-making such as color symbolism, copyright laws, and the intentional omission of restricted or sensitive areas (e.g. military bases, government facilities) or private property despite the data being available. However, the ethical and legal implications of maps created by GMS are still unclear, as is the division of responsibility for them between users, system developers, and data providers. We consider biases and ethics, which are undeniably vital, to be significant research directions on their own.

## 5 CONCLUSION

Maps serve as a powerful tool for knowledge representation and communication. By developing GMSs that lower the barriers of map-making, we can make this tool widely accessible including to the general public. This may advance the whole map-making procedure and revolutionize the way people communicate spatial knowledge. While generative AI has demonstrated its potential to generate maps, there are still technical gaps to develop a GMS effectively. In this paper, we provide a roadmap delineating the open questions and research opportunities regarding automatic and qualified map generation and customization that should be further addressed. Envisioning a future where people use and create maps as effortlessly as they do images and videos today opens up exciting possibilities for knowledge production, especially considering that maps can also represent non-geographic content, such as mind maps. Nevertheless, it is crucial to address the ethical and legal issues that accompany this technology.